\documentclass[letterpaper, 10 pt, conference]{ieeeconf}
\IEEEoverridecommandlockouts                              
\usepackage{times}
\usepackage{epsfig}
\usepackage{graphicx}
\usepackage{amsmath}
\usepackage{amssymb}
\usepackage{multirow}

\usepackage{graphicx}
\usepackage{epsfig}
\usepackage{epic,eepic}
\usepackage{times}
\usepackage{mathptmx}
\usepackage{amsfonts}
\usepackage{amsmath}
\usepackage{cite}
\usepackage{color}

\usepackage{times}

\definecolor{red}{rgb}{1,0,0}
\definecolor{green}{rgb}{0,1,0}
\definecolor{blue}{rgb}{0,0,1}
\definecolor{violet}{rgb}{1,0,1}
\definecolor{cyan}{cmyk}{1,0,0,0}
\definecolor{magenta}{cmyk}{0,1,0,0}
\definecolor{yellow}{cmyk}{0,0,1,0}

\definecolor{white}{rgb}{1,1,1}

\usepackage{jumoline}

\usepackage{arydshln}

\newcommand{\CO}[1]{}

\newcommand{\CommentOut}[1]{}

\newcommand{\noeditage}[1]{#1} \newcommand{\editage}[1]{}

\onecolumn

\begin{document}

\newcommand{\FIG}[3]{
\begin{minipage}[b]{#1cm}
\begin{center}
\includegraphics[width=#1cm]{#2}\\
{\scriptsize #3}
\end{center}
\end{minipage}
}

\newcommand{\FIGU}[3]{
\begin{minipage}[b]{#1cm}
\begin{center}
\includegraphics[width=#1cm,angle=180]{#2}\\
{\scriptsize #3}
\end{center}
\end{minipage}
}

\newcommand{\FIGm}[3]{
\begin{minipage}[b]{#1cm}
\begin{center}
\includegraphics[width=#1cm]{#2}\\
{\scriptsize #3}
\end{center}
\end{minipage}
}

\newcommand{\FIGR}[3]{
\begin{minipage}[b]{#1cm}
\begin{center}
\includegraphics[angle=-90,width=#1cm]{#2}
\\
{\scriptsize #3}
\vspace*{1mm}
\end{center}
\end{minipage}
}

\newcommand{\FIGRpng}[5]{
\begin{minipage}[b]{#1cm}
\begin{center}
\includegraphics[bb=0 0 #4 #5, angle=-90,clip,width=#1cm]{#2}\vspace*{1mm}
\\
{\scriptsize #3}
\vspace*{1mm}
\end{center}
\end{minipage}
}

\newcommand{\FIGpng}[5]{
\begin{minipage}[b]{#1cm}
\begin{center}
\includegraphics[bb=0 0 #4 #5, clip, width=#1cm]{#2}\vspace*{-1mm}\\
{\scriptsize #3}
\vspace*{1mm}
\end{center}
\end{minipage}
}

\newcommand{\FIGtpng}[5]{
\begin{minipage}[t]{#1cm}
\begin{center}
\includegraphics[bb=0 0 #4 #5, clip,width=#1cm]{#2}\vspace*{1mm}
\\
{\scriptsize #3}
\vspace*{1mm}
\end{center}
\end{minipage}
}

\newcommand{\FIGRt}[3]{
\begin{minipage}[t]{#1cm}
\begin{center}
\includegraphics[angle=-90,clip,width=#1cm]{#2}\vspace*{1mm}
\\
{\scriptsize #3}
\vspace*{1mm}
\end{center}
\end{minipage}
}

\newcommand{\FIGRm}[3]{
\begin{minipage}[b]{#1cm}
\begin{center}
\includegraphics[angle=-90,clip,width=#1cm]{#2}\vspace*{0mm}
\\
{\scriptsize #3}
\vspace*{1mm}
\end{center}
\end{minipage}
}

\newcommand{\FIGC}[5]{
\begin{minipage}[b]{#1cm}
\begin{center}
\includegraphics[width=#2cm,height=#3cm]{#4}~$\Longrightarrow$\vspace*{0mm}
\\
{\scriptsize #5}
\vspace*{8mm}
\end{center}
\end{minipage}
}

\newcommand{\FIGf}[3]{
\begin{minipage}[b]{#1cm}
\begin{center}
\fbox{\includegraphics[width=#1cm]{#2}}\vspace*{0.5mm}\\
{\scriptsize #3}
\end{center}
\end{minipage}
}

\newcommand{\acprPaperID}{25}

\newcommand{\figB}{
\begin{figure}[t]
 \begin{center}
\FIG{8}{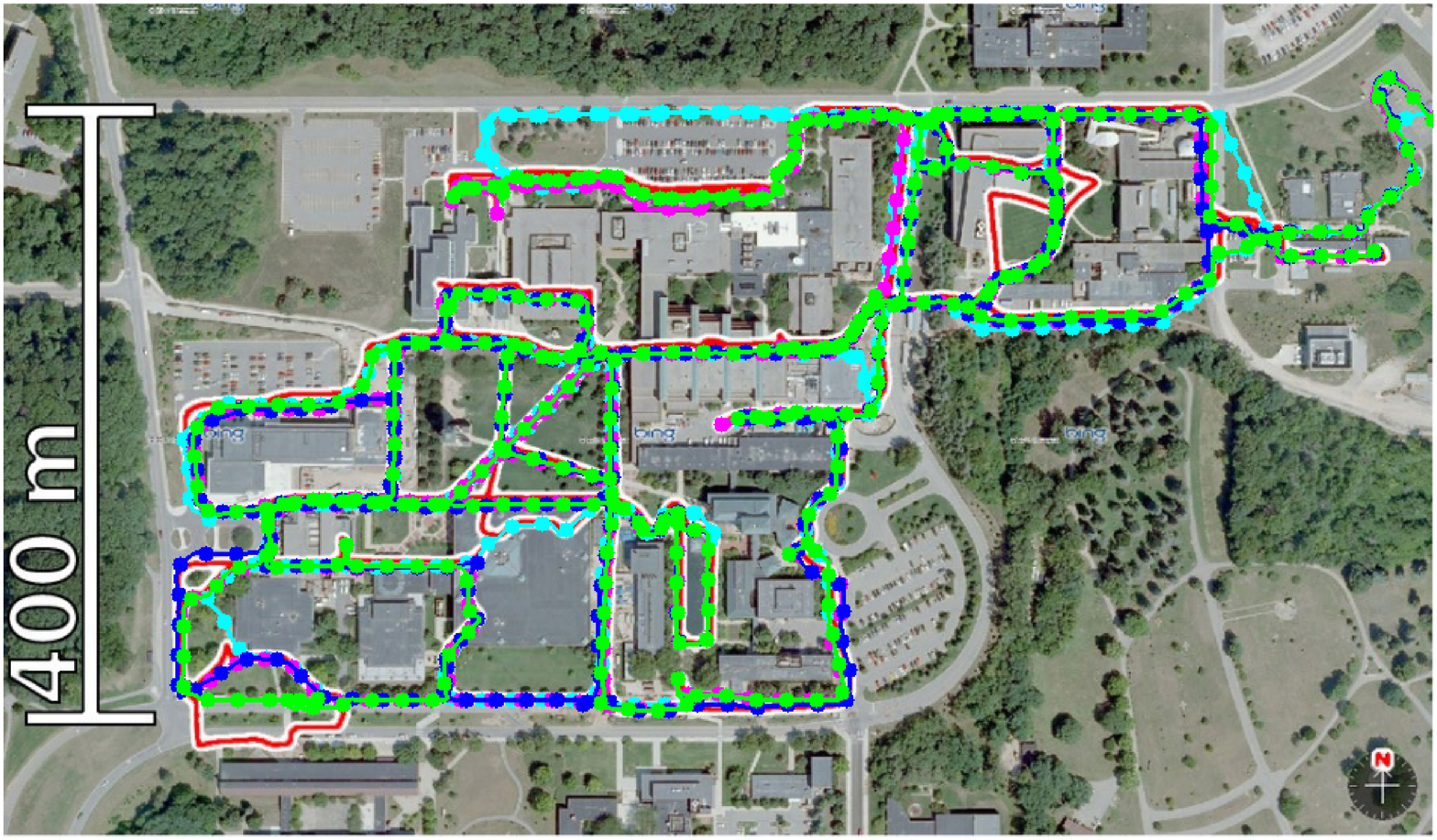}{(a)}\\
\FIG{8}{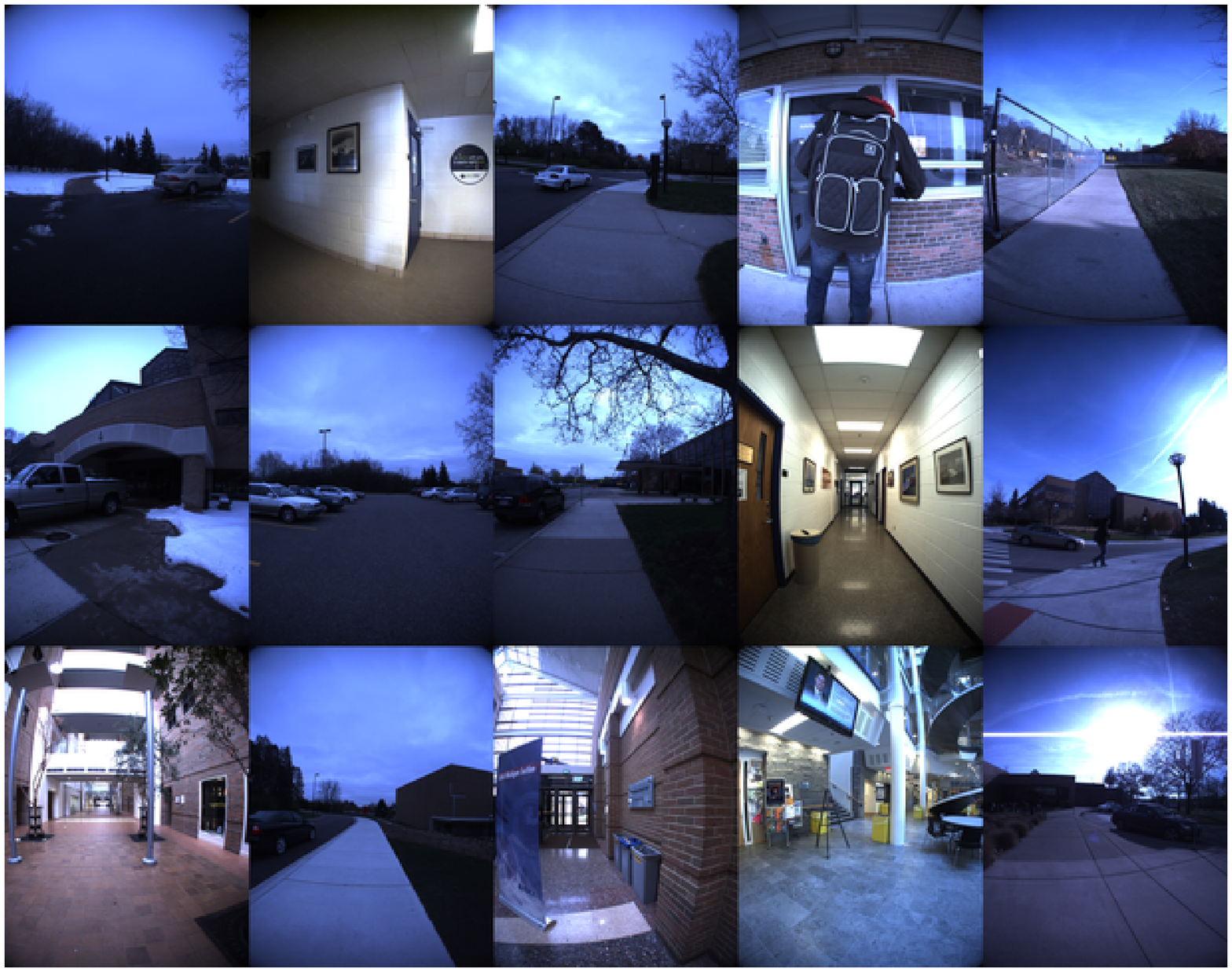}{(b)}\\
 \end{center}
 \caption{Experimental environment, (a) vehicle trajectories and (b) changing objects.} \label{fig:nclt}
\end{figure}
}

\newcommand{\tabA}{
\begin{table}[t]
\begin{center}
\caption{Top-$X$ accuracy for AE method [\%].}\label{tab:A}
\begin{tabular}{r|r|r|r|r|r|r|} \hline
\multirow{2}{*}{top-$X$} & \multicolumn{4}{|c|}{generic} & \multicolumn{2}{|c|}{season-specific} \\ \cline{2-7}
 & k=1 & k=5 & k=10 & k=15 & k=1 & k=10 \\ \hline \hline
5\% & 1.5 & 1.5 & 1.0 & 1.8 & 1.6 & 2.3 \\ \hline
10\% & 7.0 & 7.8 & 8.1 & 8.3 & 8.1 & 10.3 \\ \hline
15\% & 17.5 & 18.4 & 19.3 & 18.1 & 17.7 & 19.6 \\ \hline
20\% & 28.3 & 31.1 & 31.6 & 30.9 & 29.5 & 35.1 \\ \hline
\end{tabular}
\end{center}
\end{table}
}

\newcommand{\tabB}{
\begin{table}[t]
   \begin{center}
    \caption{Basic performance (top-$X$ accuracy [\%]).}\label{tab:B}
\begin{tabular}{r|r|r|r|r|} \hline    
\multirow{2}{*}{top-$X$} & \multicolumn{2}{|c|}{IoU$\ge$50[\%]} & \multicolumn{2}{|c|}{IoU$\ge$25[\%]} \\ \cline{2-5}    
 & LCD & AE & LCD & AE \\ \hline \hline
      5\% & 6.1 & 2.0 & 14.1 & 6.2 \\ \hline
      10\% & 11.1 & 4.6 & 27.3 & 14.4  \\ \hline
      15\% & 28.7 & 7.4 & 57.5 & 25.3 \\ \hline
      20\% & 47.0 & 11.6 & 71.4 & 38.3 \\ \hline
   \end{tabular}
  \end{center}
\end{table}
}

\newcommand{\figJ}{
\begin{figure}[t]
\begin{center}
\FIG{7}{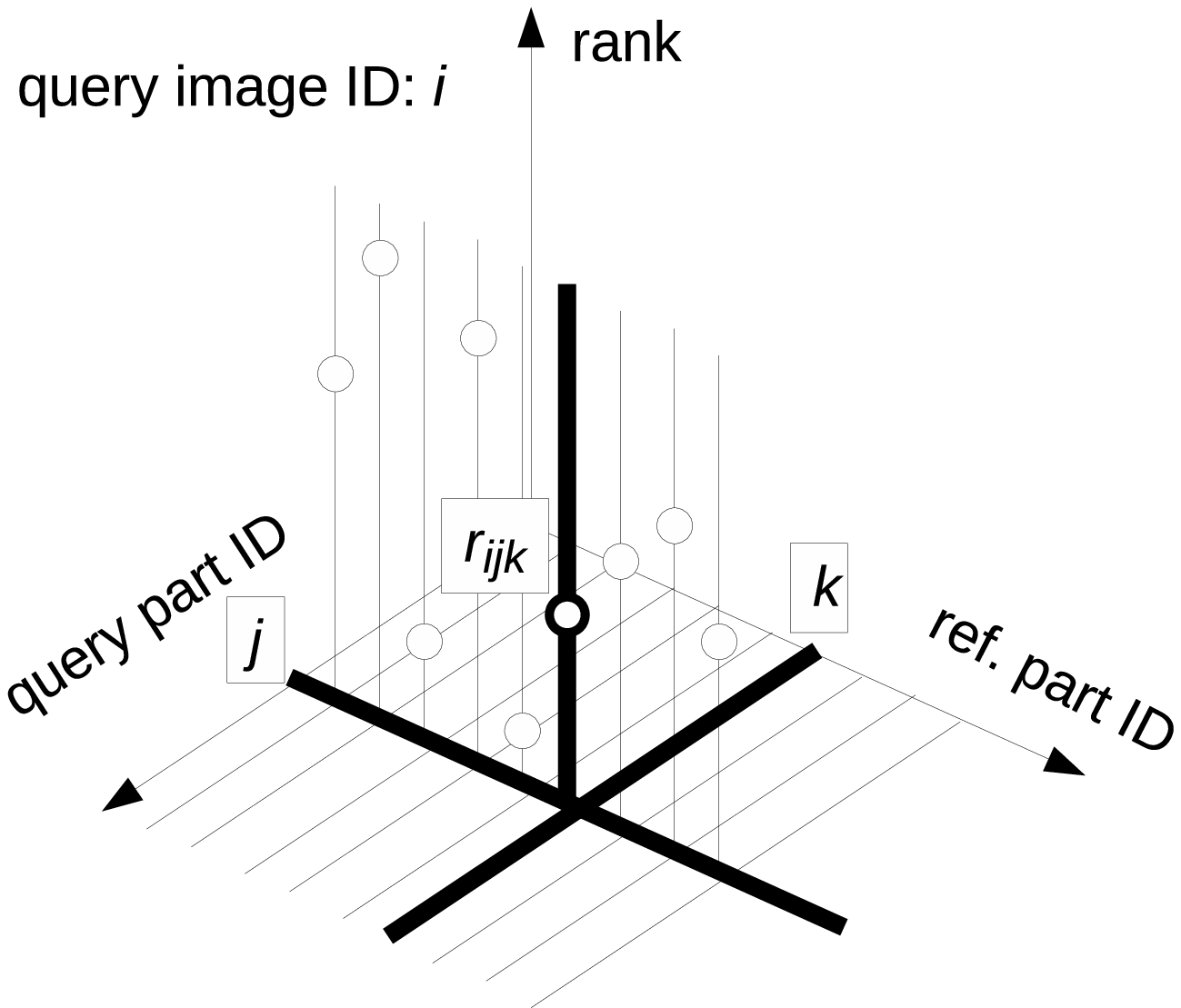}{(a) Problem formulation}\\
\FIG{7}{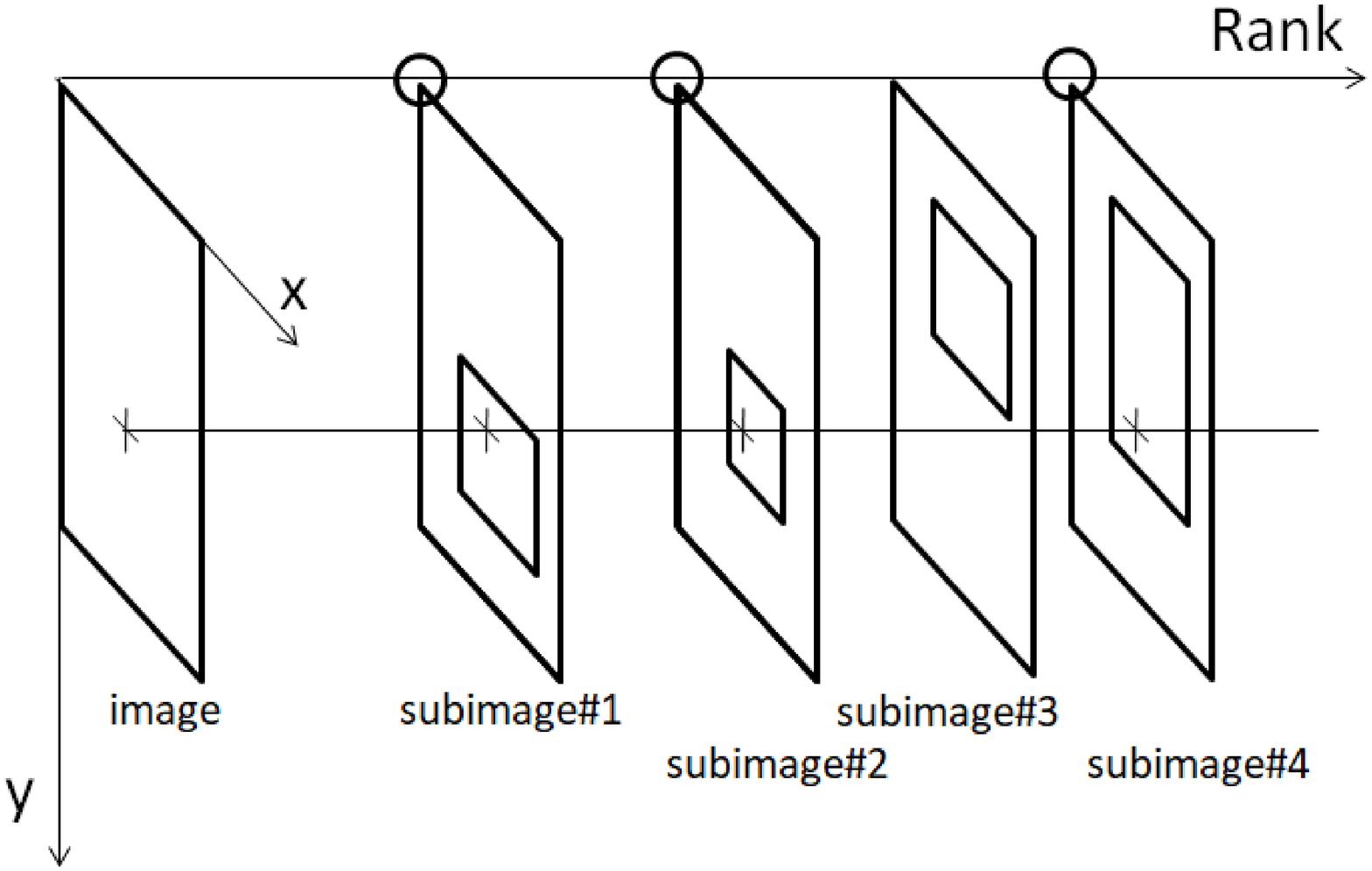}{(b) OPRs vs. rank values}\vspace*{-3mm}\\
\end{center}
\caption{%
LoC prediction.
(a)
Given an $i$-th query image,
we
perform VPR for each $j$-th scene part (OPR) in the query image
to obtain
a rank value $r_{ijk}$ for each $k$-th OPR in the ground-truth (GT) reference image.
(b)
To compute a pixel-wise LoC map,
the bounding boxes and rank values of the OPRs are
translated into a pixel-wise LoC map,
by aggregating the rank values of
the OPRs that belong to each pixel of interest (+).
}
 \label{fig:fig10}
\end{figure}
}

\newcommand{\figM}{
\begin{figure}[t]
\begin{center}
\FIG{8}{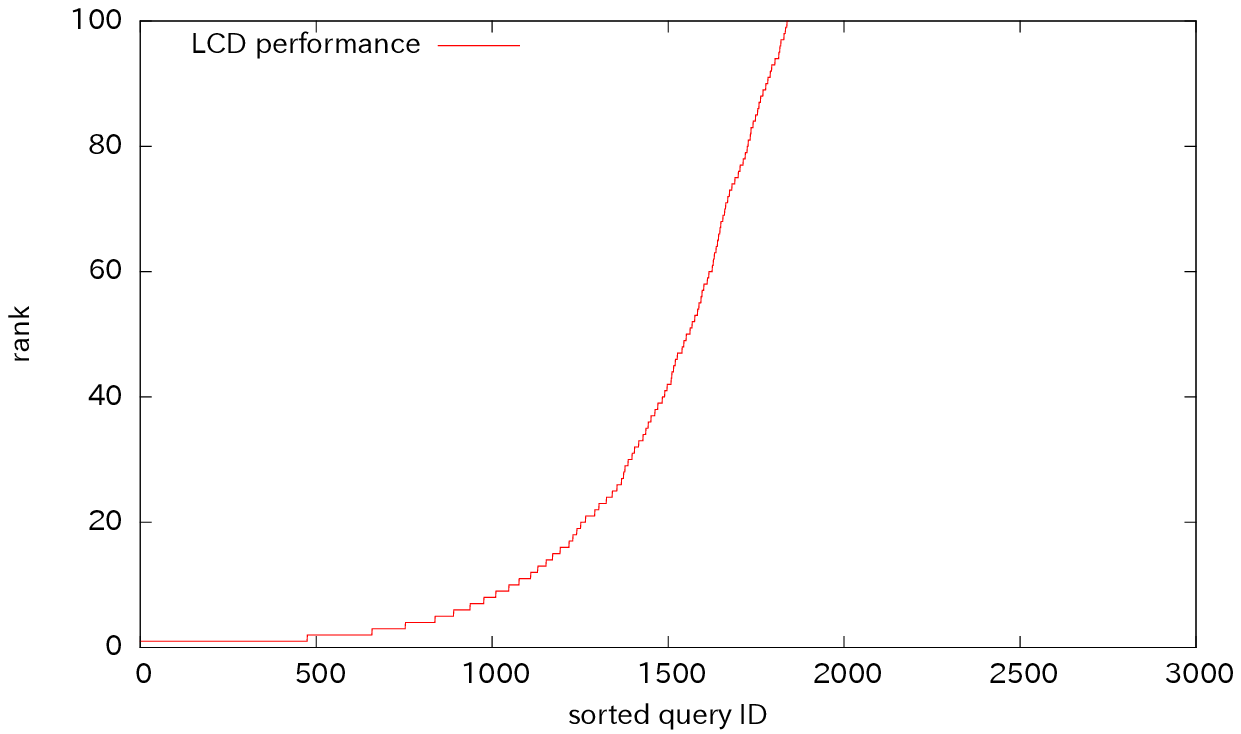}{}
\caption{LCD performance.
}\label{fig:M}
\end{center}
\end{figure}
}

\author{Tanaka Kanji ~~~~ Yamaguchi Kousuke ~~~~ Sugimoto Takuma
\thanks{Our work has been supported in part by JSPS KAKENHI Grant-in-Aid for Scientific Research (C) 26330297,
and (C) 17K00361.}
\thanks{K. Tanaka is with Faculty of Engineering of Graduate School of Engineering, University of Fukui.
K. Yamaguchi and T. Sugimoto 
are with Graduate School of Engineering, University of Fukui, Japan.
{\tt\small tnkknj@u-fukui.ac.jp}}%
\vspace*{-5mm}}
\title{\LARGE \bf
Simultaneous Detection of Loop-Closures and Changed Objects
}

\maketitle

\section*{\centering Abstract}
\textit{
Loop-closure detection (LCD) in large non-stationary environments remains an important challenge
in robotic visual simultaneous localization and mapping (vSLAM).
To reduce computational and perceptual complexity, it is helpful if a vSLAM system has the ability
to perform image change detection (ICD).
Unlike previous applications of ICD,
time-critical vSLAM applications cannot assume an offline background modeling stage,
or rely on maintenance-intensive background models.
To address this issue,
we introduce a novel maintenance-free ICD framework
that requires no background modeling.
Specifically,
we demonstrate that LCD can be reused as the main
process for ICD with minimal extra cost.
Based on these concepts,
we develop
a novel vSLAM component that enables
simultaneous
LCD and ICD.
ICD experiments based on
challenging cross-season
LCD
scenarios
validate the efficacy of the proposed method.
}

\section{Introduction}

Recent progress in robotic visual
simultaneous localization and mapping (vSLAM) has led to the development of various practical vSLAM systems (e.g., ORB-SLAM) that are able to map large non-stationary environments via robot-centric monocular vision. Loop-closure detection (LCD),
which is the problem of detecting loop-closure events,
is crucial for addressing the inherently accumulative self-localization errors in vSLAM \cite{fabmap}.
However, LCD in large non-stationary environments
(e.g., cross-season LCD \cite{kanji2015cross})
remains a significant challenge. A major source of difficulty is
the large number of possible changes between live images and
a map (e.g., car parking, furniture movement),
which grows combinatorially relative to map size.

To reduce computational and perceptual complexity, it is helpful if a vSLAM system has the ability
to detect changed objects.
The detected changes can then be used to inspect and update the corresponding
parts of a map. 

Detecting changes in a query live image
relative to a pre-trained background (or reference) model is a fundamental problem
in computer vision
called image change detection (ICD) \cite{ccr1}
that has been studied in many different contexts including remote sensing \cite{remotesensing} and surveillance \cite{surveillance}.

In these classical contexts, the problem is typically formulated as a two-stage offline-online process, where the online process aims to detect changes in a live image,
relative to an offline pre-trained background model. However, such
separate offline-online stages are not
applicable to vSLAM,
which has no offline stage.
Furthermore,
it is impractical
to maintain a background model online within the real-time budget allowed for vSLAM.

To address the above issues, we introduce a novel maintenance-free ICD algorithm
that requires no background modeling stage.
Specifically,
we propose reusing the latest map database and LCD engine,
which are continually maintained and kept up-to-date by vSLAM,
in place of a maintenance-intensive background model and detector engine, respectively.
This idea is supported by the recent findings in the field of 
multi-experience localization \cite{churchill2013experience}
that errors in self-localization (i.e., LCD) can be viewed as a good indicator of
inconsistency (i.e., changes) between encountered conditions and a map.
This approach
has two main merits. First, we can introduce such an ICD system
with very little additional cost.
Second, we can directly detect
degradation of map quality (i.e.,
the need for map updating) in terms of LCD errors.

In this paper,
we present a novel vSLAM component,
called ``LCD-ICD"
that enables
the simultaneous detection of loop-closures and changed objects.
LCD-based ICD is motivated by our previous work \cite{kanji2019detection},
but that work focused on a pose-tracking scenario,
rather than the global localization (i.e., LCD) scenario
addressed in this paper.
Note that such a simultaneous formulation is
required because localizing changed objects in the world requires
a reliable estimate of robot pose (i.e., LCD).
We employ the bag-of-local-features (BoLF) image representation
because it is a
standardized representation for state-of-the-art LCD systems \cite{Garcia-Fidalgo2018}
and an effective image model for ICD \cite{kim2005real}.
We adopt
a generic LCD formulation
that models an LCD system as a ranking function.
We also derive unsupervised rank fusion techniques from the field of multi-media information retrieval \cite{EarlyLateFusionNew}
to fuse different ranking results from different local query features
into a pixel-wise likelihood-of-change (LoC) measure.
Consequently,
our ICD system is maintenance-free, requiring no background model
or detector engine.
It is also agnostic to the choice of LCD system.
We implemented the proposed ICD framework and evaluated
the feasibility of simultaneous LCD and ICD in a practical challenging cross-season LCD scenario \cite{nclt}.
The experimental results demonstrate that the proposed LCD-ICD method is
comparable to
or superior to
standard ICD methods, despite the fact that the proposed system is maintenance-free.

\section{LCD Framework}

We follow a standard 
LCD framework consisting of three distinct stages.
At each time step, (1) the current live image is converted
into a BoLF representation (Section \ref{sec:bow})
and (2) a similarity search over a database of reference images is performed
to find the nearest neighbor reference images (Section \ref{sec:retrieval}).
Finally,
(3) the map is updated by incorporating the latest live image
for future LCD tasks (Section \ref{sec:db}).
Each of these stages is detailed in the following subsections.

\subsection{BoLF Representation}\label{sec:bow}

Converting a live image
into a BoLF representation is an important stage of the LCD process. First, semantically
coherent object proposal regions (OPRs)
representing scene parts are extracted from an image.
Next, each OPR is converted into a visual feature.
These two processes are detailed bellow.

For OPR extraction, we introduce supervised and unsupervised object proposal methods
that extract OPRs in the form of bounding boxes (BBs).
The supervised proposal method from \cite{yolo}
is employed
to extract 1-11 OPRs per image (Fig. \ref{fig:fig10}b).
The unsupervised proposal method generates five additional bounding boxes $[w/3, 2w/3]$$\times$$[h/3, 2h/3]$, $[0, 2w/3]$$\times$$[0, 2h/3]$, $[w/3, w]$$\times$$[0, 2h/3]$, $[0, 2w/3]$$\times$$[h/3, h]$, and $[w/3, w]$$\times$$[h/3, h]$ for a $w\times h$ image.
Therefore,
we extract a total of 6-16 OPRs per image.

For feature extraction,
we
use the intermediate layer
of an unsupervised AE
as a feature extractor.
This design choice was motivated by the recent success
of AEs in LCD applications \cite{aelcd}.
Importantly,
it can be trained in an unsupervised manner,
requiring
only unlabeled images 
from the target environments.
For these properties, future widespread use of such AE-based
LCD systems are expected. 
The authors of
\cite{aelcd}
trained a domain-specific AE
using training images
that were collected
from the target environment.
In contrast,
we pretrain a generic AE using an independent train image set, with the
goal of performing domain-generic LCD.
Our AE architecture uses a layer structure of 
128-64-32-16-16-32-64-128 nodes,
and
is trained on
a collection of reference images.
Every image is resized to 256$\times$256
pixels
prior to inputting it into the AE.

\subsection{Similarity Search}\label{sec:retrieval}

The similarity search stage aims to find similar images to a query image.
The BoLFs in all reference images are ranked in descending order of similarity.
In this work,
the L2 norm
was used as a similarity measure.

\subsection{Map Updating}\label{sec:db}

The map updating stage aims to insert
the current live image into the map database
for future LCD.
Each local feature in the live image
is used as an index
into the map database.

\figJ

\section{ICD based on LCD}\label{sec:icdbylcd}

We now introduce our ICD component
as a function of LCD.
In this section,
we begin
by assuming the availability of a ``ground-truth (GT)" reference image
as the reference image whose viewpoint is closest to the current viewpoint.
This assumption will be relaxed in the following section.
Fig. \ref{fig:fig10} illustrates the formulation of our LoC prediction problem.

The majority of works on BoLF based LCD models use LCD as a ranking function. 
Let us assume that a total of $n$ images are contained in a map database $D=\{I_i\}_{i=1}^n$. Each image $I_i$ has a set of local features $\{x_{ij}\}_{j=1}^{d_i}$, where $d_i$ is the number of local features. Given a query image,
a similarity search (section \ref{sec:retrieval})
is performed using each local feature as a query. Then, for each local feature in the query image,
a ranked list of reference images in the descending order of similarity is returned.

In our approach, this LCD system is reused to estimate the LoC of a given query image (Fig. \ref{fig:fig10}a). Given the local features $\{x_{ij}\}_{j=1}^{d_i}$ 
of multiple overlapping OPRs in a query image, we retrieve a map and obtain $d_i$ rank lists.
Then, 
the LoC of each $j$-th OPR is measured 
based on the rank values
$\{\{r_{ijk}\}_{k=1}\}_{j=1}^{d_i}$
(ascending order of similarity)
of 
the $k$-th OPR in
the GT reference image
within the $j$-th ranked list.

We compute the anomaly score
of the $j$-th OPR in
the $i$-th query image
as the minimum of the related reference OPRs:
\begin{equation}
r_{ij} = \min_{k} r_{ijk}.
\end{equation}

Now we will characterize LoC measures for the LCD method (Fig. \ref{fig:fig10}b). As mentioned above,
rank values are obtained for each $j$-th
OPR in the $i$-th query image, in the form of rank values of
GT reference images.
Therefore,
it is natural to apply a pooling technique to aggregate
these values (i.e., OPR-level LoC values) into a pixel-wise LoC value.

The rank aggregation problem
was explored in the
context of
part-based self-localization
in our previous study \cite{kanji2015unsupervised}.
The method used in this study 
is based on
our previous method
with a few key modifications:
First, our previous study aimed at image-level ranking,
whereas this study aims to obtain pixel-level rank values.
Second, the previous method took
non-overlapping query subimages (from color-based segmentation)
as inputs,
whereas the current
method takes
overlapping query subimages (unsupervised/supervised OPRs)
with {\it variable}
amounts of overlap per pixel.

To address this issue,
we must perform the novel task of pixel-wise rank fusion \cite{kanji2019detection}. 
Formally,
we adopt
the recently presented extension of 
{\it variable}
length rank lists
for MMR \cite{mmr15},
and fuse per-pixel ranking results as follows:
\begin{equation}
r_{i}[p]=|J[p]|/\left(\sum_{j\in J[p]} r_{ij}[p]^{-1} \right), \label{eqn:1}
\end{equation}
where
$J[p]$
is the set of identifiers of OPRs belonging to pixel $p$.

\section{Simultaneous LCD and ICD}\label{sec:simultaneous}

We now consider a more practical scenario
in which
GT references are not available.
Instead, a collection of $Y$ viewpoint hypotheses are provided by the similarity search
of LCD (section \ref{sec:retrieval}).

In our experimental system, the similarity search of LCD is based on the naive Bayes nearest neighbor (NBNN) similarity metric,
which was proven to be effective in our previous study \cite{NbnnIros16}.
Given a query image BoLF $f_i$ and
a collection of reference image BoLFs $\{f_{i'}^j\}$, the problem of finding
the nearest neighbor reference image $j$ to a query image is
defined as follows:
\begin{equation}
j = \arg \min_{j} \frac{1}{n_i} \sum_i^{n_i} \min_{i'} | f_i - f_{i'}^j |,
\end{equation}
where $n_i$ is the number of local features in the $i$-th image.

We consider
the $Y$ top-ranked viewpoint hypotheses (e.g., $Y=10$)
top-ranked
from the similarity search
and aggregate their 
LoC images
into a single final LoC image.

For aggregation,
we consider the nature of
the proposed LCD-ICD system.
Incorrect viewpoint hypotheses
represent large LCD errors
and 
cause an increase in
the LoC value
at every pixel in the LoC image.
In other words,
the LoC value
at a pixel
is much higher
for incorrect viewpoint hypotheses
than
it is for correct viewpoint hypotheses.
To filter out
unreliably large LoC values,
we perform
pixel-wise
minimum pooling
to aggregate LoC image hypotheses 
in the form of
$r_i[p]=min_h r_i[p][h]$,
where
$r_i[p][h]$
is the ranking result
at pixel $p$ for
the
$h$-th viewpoint hypothesis.

\figB

\section{Experiments}

We
evaluated
the proposed ICD framework
on
a challenging
cross-season
LCD
scenario.

\subsection{Settings}\label{sec:settings}

We used a public dataset, North Campus long-term (NCLT) dataset
for our experiments \cite{nclt}. The NCLT dataset is a large-scale, long-term autonomy dataset for robotics research
that was collected at the University of Michigan's North Campus
by a Segway vehicle platform. The data we used
in this study includes view image sequences along the vehicle trajectories acquired by the front facing camera of the Ladybug3 platform (Fig. \ref{fig:nclt}).
From the
perspective of ICD benchmarking,
the NCLT dataset has desirable properties.
It includes
various types
of changing images
such as cars, pedestrians, building construction, construction machines, posters, tables and whiteboards with wheels,
from seamless indoor and outdoor navigations of the Segway vehicle.
Additionally,
it
has recently
gained significant popularity
as a benchmark in
the SLAM community  \cite{jmangelson-2018a}.

\noeditage{
}

In this study, we used four datasets
labeled
``2012/1/22", ``2012/3/31", ``2012/8/4", and ``2012/11/17"
(denoted WI, SP, SU, and AU, respectively)
collected
from four different seasons.
We annotated 986 different changing objects
with bounding boxes in total.
The annotations are found in all 12 possible pairings of query and reference seasons.
The image size is 1232$\times$1616.
Additionally,
we prepared a collection of 1,973 random destructor images,
that
are independent
of the 986 annotated images
and do not contain changing objects.
We then merged
the 1,973 destructor images
and 986 annotated images to obtain
a map database containing
2,959 images.
Fig.\ref{fig:nclt}b
presents
examples of changing objects
in the dataset.

Performance on
the ICD task
is evaluated in terms of
top-$X,Y$ accuracy.
First,
we estimate
an LoC image
using an ICD algorithm
on the top-$Y$
self-localization hypotheses.
We then impose a 2D grid with $10\times 10$ pixel sized cells
on the query image
and estimate an LoC for each cell by max-pooling the pixel-wise LoC values from
all pixels 
that belong to that cell.
Next, all cells from all images in the map are sorted
in descending order of LoC,
and the
accuracies of the top-$X$ items in the list are evaluated.
We evaluate the top-$X,Y$ accuracy
for different $X$ thresholds in
consideration
for the intersection-over-union (IoU) criterion \cite{yolo}.
For a specific $X$ threshold,
a successful detection is defined as a changed object whose annotated bounding box is sufficiently covered (IoU $\ge$ 50\% or 25\%) by the top-$X$ percent of cells.

\subsection{Comparing Methods}\label{sec:methods}

We
compared the proposed method
to
a benchmark method,
namely
AE reconstruction error -based anomaly detection (denoted AE) \cite{aeicd}.

The AE method evaluates the LoC at each pixel
based on the L2 distance between
that pixel and a pixel reconstructed by an AE.
Every image is resized to 128$\times$128 pixels
prior to inputting it to AE.
To avoid overfitting,
we constructed $k$ different AEs.
First,
the reference image set
was
divided into $k$ disjoint clusters,
and each AE was trained
on one cluster.
For set division,
we employed the k-means clustering algorithm
with the number of clusters set to $k$.
Each trained AE was used as the 
background model
for the reference images belonging to the corresponding cluster.

\tabA
\tabB

Table \ref{tab:A}
lists
the results of preliminary experiments
on the influence of $k$ on
top-$X$ accuracy when
assuming the availability of a GT viewpoint.
The $k$ AEs were trained on a single season-generic training set (generic),
or four season-specific training sets (season-specific).
In the experiments described bellow,
we set $k$=10
and used the
season-generic AEs.

\figM

\subsection{Basic Performance}\label{sec:smallexp}

In the current paper,
we focused on the basic performance of
the LCD-ICD framework
and compared its performance to 
the benchmark method.
To this end,
we considered a relatively small map database,
and constructed
a database with $N=100$ images
consisting
of a GT reference image
and
$(N-1)$ non-GT reference images,
which were random samples from the 2,959
total reference images.

We employed
an LCD system
that represents each image
with a size-six BoLF
consisting of
a full image AE feature
and 
five unsupervised OPR AE features (section \ref{sec:bow}).
Following our previous work in \cite{kanji2015unsupervised},
relative contribution of the NBNN value of each OPR
compared to that of the entire image
was set 1/20.
We set the number of
viewpoint hypotheses
$Y$
to 10.
This means that
for ICD,
each query image was compared against
10 different reference images
retrieved from the map.
Fig. \ref{fig:M}
presents the 
performance results for the self-localization problem.

Table \ref{tab:B}
lists the 
top-$X,Y$ accuracies
of 
the proposed method (LCD)
and the benchmark method.
One can see that
the performance of the proposed method is much better than
that of the AE method.

\section{Conclusions}

The primary contribution of this paper is
the proposal of a novel vSLAM component
called
LCD-ICD
that facilitates
simultaneous
LCD and ICD.
We 
demonstrated
that LCD can be reused as the main
processing
for ICD with minimal extra cost.
We also designed
LCD-agnostic strategies
for
fusing information from multiple local features
and
aggregating LoC images from different viewpoint hypotheses.
The result is
a maintenance-free
LCD-ICD framework
that
requires no background modeling
or detector engine
and is
agnostic to the choice of LCD systems.

\bibliographystyle{IEEEtran}
\bibliography{cite}

\end{document}